# Solving Multiclass Learning Problems via
# Error-Correcting Output Codes

**Thomas G. Dietterich**                                    TGD@CS.ORST.EDU
*Department of Computer Science, 303 Dearborn Hall*
*Oregon State University*
*Corvallis, OR 97331 USA*

**Ghulum Bakiri**                                          EB004@ISA.CC.UOB.BH
*Department of Computer Science*
*University of Bahrain*
*Isa Town, Bahrain*

## Abstract

Multiclass learning problems involve finding a definition for an unknown function $f(\mathbf{x})$ whose range is a discrete set containing $k > 2$ values (i.e., $k$ "classes"). The definition is acquired by studying collections of training examples of the form $\langle \mathbf{x}_i, f(\mathbf{x}_i) \rangle$. Existing approaches to multiclass learning problems include direct application of multiclass algorithms such as the decision-tree algorithms C4.5 and CART, application of binary concept learning algorithms to learn individual binary functions for each of the $k$ classes, and application of binary concept learning algorithms with distributed output representations. This paper compares these three approaches to a new technique in which error-correcting codes are employed as a distributed output representation. We show that these output representations improve the generalization performance of both C4.5 and backpropagation on a wide range of multiclass learning tasks. We also demonstrate that this approach is robust with respect to changes in the size of the training sample, the assignment of distributed representations to particular classes, and the application of overfitting avoidance techniques such as decision-tree pruning. Finally, we show that—like the other methods—the error-correcting code technique can provide reliable class probability estimates. Taken together, these results demonstrate that error-correcting output codes provide a general-purpose method for improving the performance of inductive learning programs on multiclass problems.

## 1. Introduction

The task of learning from examples is to find an approximate definition for an unknown function $f(\mathbf{x})$ given training examples of the form $\langle \mathbf{x}_i, f(\mathbf{x}_i) \rangle$. For cases in which $f$ takes only the values $\{0, 1\}$—binary functions—there are many algorithms available. For example, the decision-tree methods, such as C4.5 (Quinlan, 1993) and CART (Breiman, Friedman, Olshen, & Stone, 1984) can construct trees whose leaves are labeled with binary values. Most artificial neural network algorithms, such as the perceptron algorithm (Rosenblatt, 1958) and the error backpropagation (BP) algorithm (Rumelhart, Hinton, & Williams, 1986), are best suited to learning binary functions. Theoretical studies of learning have focused almost entirely on learning binary functions (Valiant, 1984; Natarajan, 1991).

In many real-world learning tasks, however, the unknown function $f$ often takes values from a discrete set of "classes": $\{c_1, \ldots, c_k\}$. For example, in medical diagnosis, the function might map a description of a patient to one of $k$ possible diseases. In digit recognition (e.g.,

LeCun, Boser, Denker, Henderson, Howard, Hubbard, & Jackel, 1989), the function maps each hand-printed digit to one of $k = 10$ classes. Phoneme recognition systems (e.g., Waibel, Hanazawa, Hinton, Shikano, & Lang, 1989) typically classify a speech segment into one of 50 to 60 phonemes.

Decision-tree algorithms can be easily generalized to handle these "multiclass" learning tasks. Each leaf of the decision tree can be labeled with one of the $k$ classes, and internal nodes can be selected to discriminate among these classes. We will call this the *direct multiclass* approach.

Connectionist algorithms are more difficult to apply to multiclass problems. The standard approach is to learn $k$ individual binary functions $f_1, \ldots, f_k$, one for each class. To assign a new case, $\mathbf{x}$, to one of these classes, each of the $f_i$ is evaluated on $\mathbf{x}$, and $\mathbf{x}$ is assigned the class $j$ of the function $f_j$ that returns the highest activation (Nilsson, 1965). We will call this the *one-per-class* approach, since one binary function is learned for each class.

An alternative approach explored by some researchers is to employ a *distributed output code*. This approach was pioneered by Sejnowski and Rosenberg (1987) in their widely-known NETtalk system. Each class is assigned a unique binary string of length $n$; we will refer to these strings as "codewords." Then $n$ binary functions are learned, one for each bit position in these binary strings. During training for an example from class $i$, the desired outputs of these $n$ binary functions are specified by the codeword for class $i$. With artificial neural networks, these $n$ functions can be implemented by the $n$ output units of a single network.

New values of $\mathbf{x}$ are classified by evaluating each of the $n$ binary functions to generate an $n$-bit string $s$. This string is then compared to each of the $k$ codewords, and $\mathbf{x}$ is assigned to the class whose codeword is closest, according to some distance measure, to the generated string $s$.

As an example, consider Table 1, which shows a six-bit distributed code for a ten-class digit-recognition problem. Notice that each row is distinct, so that each class has a unique codeword. As in most applications of distributed output codes, the bit positions (columns) have been chosen to be meaningful. Table 2 gives the meanings for the six columns. During learning, one binary function will be learned for each column. Notice that each column is also distinct and that each binary function to be learned is a disjunction of the original classes. For example, $f_{vl}(\mathbf{x}) = 1$ if $f(\mathbf{x})$ is 1, 4, or 5.

To classify a new hand-printed digit, $\mathbf{x}$, the six functions $f_{vl}, f_{hl}, f_{dl}, f_{cc}, f_{ol}$, and $f_{or}$ are evaluated to obtain a six-bit string, such as 110001. Then the distance of this string to each of the ten codewords is computed. The nearest codeword, according to Hamming distance (which counts the number of bits that differ), is 110000, which corresponds to class 4. Hence, this predicts that $f(\mathbf{x}) = 4$.

This process of mapping the output string to the nearest codeword is identical to the decoding step for error-correcting codes (Bose & Ray-Chaudhuri, 1960; Hocquenghem, 1959). This suggests that there might be some advantage to employing error-correcting codes as a distributed representation. Indeed, the idea of employing error-correcting, distributed representations can be traced to early research in machine learning (Duda, Machanik, & Singleton, 1963).

Table 1: A distributed code for the digit recognition task.

| Class | Code Word | | | | | |
|-------|-----|-----|-----|-----|-----|-----|
|       | vl  | hl  | dl  | cc  | ol  | or  |
| 0     | 0   | 0   | 0   | 1   | 0   | 0   |
| 1     | 1   | 0   | 0   | 0   | 0   | 0   |
| 2     | 0   | 1   | 1   | 0   | 1   | 0   |
| 3     | 0   | 0   | 0   | 0   | 1   | 0   |
| 4     | 1   | 1   | 0   | 0   | 0   | 0   |
| 5     | 1   | 1   | 0   | 0   | 1   | 0   |
| 6     | 0   | 0   | 1   | 1   | 0   | 1   |
| 7     | 0   | 0   | 1   | 0   | 0   | 0   |
| 8     | 0   | 0   | 0   | 1   | 0   | 0   |
| 9     | 0   | 0   | 1   | 1   | 0   | 0   |

Table 2: Meanings of the six columns for the code in Table 1.

| Column position | Abbreviation | Meaning |
|-----------------|--------------|---------|
| 1 | vl | contains vertical line |
| 2 | hl | contains horizontal line |
| 3 | dl | contains diagonal line |
| 4 | cc | contains closed curve |
| 5 | ol | contains curve open to left |
| 6 | or | contains curve open to right |

Table 3: A 15-bit error-correcting output code for a ten-class problem.

| Class | Code Word | | | | | | | | | | | | | | |
|-------|-------|-------|-------|-------|-------|-------|-------|-------|-------|-------|----------|----------|----------|----------|----------|
|       | $f_0$ | $f_1$ | $f_2$ | $f_3$ | $f_4$ | $f_5$ | $f_6$ | $f_7$ | $f_8$ | $f_9$ | $f_{10}$ | $f_{11}$ | $f_{12}$ | $f_{13}$ | $f_{14}$ |
| 0 | 1 | 1 | 0 | 0 | 0 | 0 | 1 | 0 | 1 | 0 | 0 | 1 | 1 | 0 | 1 |
| 1 | 0 | 0 | 1 | 1 | 1 | 1 | 0 | 1 | 0 | 1 | 1 | 0 | 0 | 1 | 0 |
| 2 | 1 | 0 | 0 | 1 | 0 | 0 | 0 | 1 | 1 | 1 | 1 | 0 | 1 | 0 | 1 |
| 3 | 0 | 0 | 1 | 1 | 0 | 1 | 1 | 1 | 0 | 0 | 0 | 0 | 1 | 0 | 1 |
| 4 | 1 | 1 | 1 | 0 | 1 | 0 | 1 | 1 | 0 | 0 | 1 | 0 | 0 | 0 | 1 |
| 5 | 0 | 1 | 0 | 0 | 1 | 1 | 0 | 1 | 1 | 1 | 0 | 0 | 0 | 0 | 1 |
| 6 | 1 | 0 | 1 | 1 | 1 | 0 | 0 | 0 | 0 | 1 | 0 | 1 | 0 | 0 | 1 |
| 7 | 0 | 0 | 0 | 1 | 1 | 1 | 1 | 0 | 1 | 0 | 1 | 1 | 0 | 0 | 1 |
| 8 | 1 | 1 | 0 | 1 | 0 | 1 | 1 | 0 | 0 | 1 | 0 | 0 | 0 | 1 | 1 |
| 9 | 0 | 1 | 1 | 1 | 0 | 0 | 0 | 0 | 0 | 1 | 0 | 1 | 0 | 0 | 1 |

Table 3 shows a 15-bit error-correcting code for the digit-recognition task. Each class is represented by a code word drawn from an error-correcting code. As with the distributed encoding of Table 1, a separate boolean function is learned for each bit position of the error-correcting code. To classify a new example $x$, each of the learned functions $f_0(x), \ldots, f_{14}(x)$ is evaluated to produce a 15-bit string. This is then mapped to the nearest of the ten codewords. This code can correct up to three errors out of the 15 bits.

This error-correcting code approach suggests that we view machine learning as a kind of communications problem in which the identity of the correct output class for a new example is being "transmitted" over a channel. The channel consists of the input features, the training examples, and the learning algorithm. Because of errors introduced by the finite training sample, poor choice of input features, and flaws in the learning process, the class information is corrupted. By encoding the class in an error-correcting code and "transmitting" each bit separately (i.e., via a separate run of the learning algorithm), the system may be able to recover from the errors.

This perspective further suggests that the one-per-class and "meaningful" distributed output approaches will be inferior, because their output representations do not constitute robust error-correcting codes. A measure of the quality of an error-correcting code is the minimum Hamming distance between any pair of code words. If the minimum Hamming distance is $d$, then the code can correct at least $\lfloor \frac{d-1}{2} \rfloor$ single bit errors. This is because each single bit error moves us one unit away from the true codeword (in Hamming distance). If we make only $\lfloor \frac{d-1}{2} \rfloor$ errors, the nearest codeword will still be the correct codeword. (The code of Table 3 has minimum Hamming distance seven and hence it can correct errors in any three bit positions.) The Hamming distance between any two codewords in the one-per-class code is two, so the one-per-class encoding of the $k$ output classes cannot correct any errors.

The minimum Hamming distance between pairs of codewords in a "meaningful" distributed representation tends to be very low. For example, in Table 1, the Hamming distance between the codewords for classes 4 and 5 is only one. In these kinds of codes, new columns are often introduced to discriminate between only two classes. Those two classes will therefore differ only in one bit position, so the Hamming distance between their output representations will be one. This is also true of the distributed representation developed by Sejnowski and Rosenberg (1987) in the NETtalk task.

In this paper, we compare the performance of the error-correcting code approach to the three existing approaches: the direct multiclass method (using decision trees), the one-per-class method, and (in the NETtalk task only) the meaningful distributed output representation approach. We show that error-correcting codes produce uniformly better generalization performance across a variety of multiclass domains for both the C4.5 decision-tree learning algorithm and the backpropagation neural network learning algorithm. We then report a series of experiments designed to assess the robustness of the error-correcting code approach to various changes in the learning task: length of the code, size of the training set, assignment of codewords to classes, and decision-tree pruning. Finally, we show that the error-correcting code approach can produce reliable class probability estimates.

The paper concludes with a discussion of the open questions raised by these results. Chief among these questions is the issue of why the errors being made in the different bit positions of the output are somewhat independent of one another. Without this indepen-dence, the error-correcting output code method would fail. We address this question—for the case of decision-tree algorithms—in a companion paper (Kong & Dietterich, 1995).

Table 4: Data sets employed in the study.

| Name | Number of Features | Number of Classes | Number of Training Examples | Number of Test Examples |
|---|---|---|---|---|
| glass | 9 | 6 | 214 | 10-fold xval |
| vowel | 10 | 11 | 528 | 462 |
| POS | 30 | 12 | 3,060 | 10-fold xval |
| soybean | 35 | 19 | 307 | 376 |
| audiologyS | 69 | 24 | 200 | 26 |
| ISOLET | 617 | 26 | 6,238 | 1,559 |
| letter | 16 | 26 | 16,000 | 4,000 |
| NETtalk | 203 | 54 phonemes 6 stresses | 1000 words = 7,229 letters | 1000 words = 7,242 letters |

## 2. Methods

This section describes the data sets and learning algorithms employed in this study. It also discusses the issues involved in the design of error-correcting codes and describes four algorithms for code design. The section concludes with a brief description of the methods applied to make classification decisions and evaluate performance on independent test sets.

### 2.1 Data Sets

Table 4 summarizes the data sets employed in the study. The glass, vowel, soybean, audiologyS, ISOLET, letter, and NETtalk data sets are available from the Irvine Repository of machine learning databases (Murphy & Aha, 1994).[1] The POS (part of speech) data set was provided by C. Cardie (personal communication); an earlier version of the data set was described by Cardie (1993). We did not use the entire NETtalk data set, which consists of a dictionary of 20,003 words and their pronunciations. Instead, to make the experiments feasible, we chose a training set of 1000 words and a disjoint test set of 1000 words at random from the NETtalk dictionary. In this paper, we focus on the percentage of letters pronounced correctly (rather than whole words). To pronounce a letter, both the phoneme and stress of the letter must be determined. Although there are $54 \times 6$ syntactically possible combinations of phonemes and stresses, only 140 of these appear in the training and test sets we selected.

---

1. The repository refers to the soybean data set as "soybean-large", the "audiologyS" data set as "audiology.standardized", and the "letter" data set as "letter-recognition".

## 2.2 Learning Algorithms

We employed two general classes of learning methods: algorithms for learning decision trees and algorithms for learning feed-forward networks of sigmoidal units (artificial neural networks). For decision trees, we performed all of our experiments using C4.5, Release 1, which is an older (but substantially identical) version of the program described in Quinlan (1993). We have made several changes to C4.5 to support distributed output representations, but these have not affected the tree-growing part of the algorithm. For pruning, the confidence factor was set to 0.25. C4.5 contains a facility for creating "soft thresholds" for continuous features. We found experimentally that this improved the quality of the class probability estimates produced by the algorithm in the "glass", "vowel", and "ISOLET" domains, so the results reported for those domains were computed using soft thresholds.

For neural networks, we employed two implementations. In most domains, we used the extremely fast backpropagation implementation provided by the CNAPS neurocomputer (Adaptive Solutions, 1992). This performs simple gradient descent with a fixed learning rate. The gradient is updated after presenting each training example; no momentum term was employed. A potential limitation of the CNAPS is that inputs are only represented to eight bits of accuracy, and weights are only represented to 16 bits of accuracy. Weight update arithmetic does not round, but instead performs jamming (i.e., forcing the lowest order bit to 1 when low order bits are lost due to shifting or multiplication). On the speech recognition, letter recognition, and vowel data sets, we employed the `opt` system distributed by Oregon Graduate Institute (Barnard & Cole, 1989). This implements the conjugate gradient algorithm and updates the gradient after each complete pass through the training examples (known as per-epoch updating). No learning rate is required for this approach.

Both the CNAPS and `opt` attempt to minimize the squared error between the computed and desired outputs of the network. Many researchers have employed other error measures, particularly cross-entropy (Hinton, 1989) and classification figure-of-merit (CFM, Hampshire II & Waibel, 1990). Many researchers also advocate using a softmax normalizing layer at the outputs of the network (Bridle, 1990). While each of these configurations has good theoretical support, Richard and Lippmann (1991) report that squared error works just as well as these other measures in producing accurate posterior probability estimates. Furthermore, cross-entropy and CFM tend to overfit more easily than squared error (Lippmann, personal communication; Weigend, 1993). We chose to minimize squared error because this is what the CNAPS and opt systems implement.

With either neural network algorithm, several parameters must be chosen by the user. For the CNAPS, we must select the learning rate, the initial random seed, the number of hidden units, and the stopping criteria. We selected these to optimize performance on a validation set, following the methodology of Lang, Hinton, and Waibel (1990). The training set is subdivided into a subtraining set and a validation set. While training on the subtraining set, we observed generalization performance on the validation set to determine the optimal settings of learning rate and network size and the best point at which to stop training. The training set mean squared error at that stopping point is computed, and training is then performed on the entire training set using the chosen parameters and stopping at the indicated mean squared error. Finally, we measure network performance on the test set.

For most of the data sets, this procedure worked very well. However, for the letter recognition data set, it was clearly choosing poor stopping points for the full training set. To overcome this problem, we employed a slightly different procedure to determine the stopping epoch. We trained on a series of progressively larger training sets (all of which were subsets of the final training set). Using a validation set, we determined the best stopping epoch on each of these training sets. We then extrapolated from these training sets to predict the best stopping epoch on the full training set.

For the "glass" and "POS" data sets, we employed ten-fold cross-validation to assess generalization performance. We chose training parameters based on only one "fold" of the ten-fold cross-validation. This creates some test set contamination, since examples in the validation set data of one fold are in the test set data of other folds. However, we found that there was little or no overfitting, so the validation set had little effect on the choice of parameters or stopping points.

The other data sets all come with designated test sets, which we employed to measure generalization performance.

## 2.3 Error-Correcting Code Design

We define an error-correcting code to be a matrix of binary values such as the matrix shown in Table 3. The length of a code is the number of columns in the code. The number of rows in the code is equal to the number of classes in the multiclass learning problem. A "codeword" is a row in the code.

A good error-correcting output code for a $k$-class problem should satisfy two properties:

- **Row separation.** Each codeword should be well-separated in Hamming distance from each of the other codewords.

- **Column separation.** Each bit-position function $f_i$ should be uncorrelated with the functions to be learned for the other bit positions $f_j, j \neq i$. This can be achieved by insisting that the Hamming distance between column $i$ and each of the other columns be large and that the Hamming distance between column $i$ and the *complement* of each of the other columns also be large.

The power of a code to correct errors is directly related to the row separation, as discussed above. The purpose of the column separation condition is less obvious. If two columns $i$ and $j$ are similar or identical, then when a deterministic learning algorithm such as C4.5 is applied to learn $f_i$ and $f_j$, it will make similar (correlated) mistakes. Error-correcting codes only succeed if the errors made in the individual bit positions are relatively uncorrelated, so that the number of simultaneous errors in many bit positions is small. If there are many simultaneous errors, the error-correcting code will not be able to correct them (Peterson & Weldon, 1972).

The errors in columns $i$ and $j$ will also be highly correlated if the bits in those columns are complementary. This is because algorithms such as C4.5 and backpropagation treat a class and its complement symmetrically. C4.5 will construct identical decision trees if the 0-class and 1-class are interchanged. The maximum Hamming distance between two columns is attained when the columns are complements. Hence, the column separation condition attempts to ensure that columns are neither identical nor complementary.

Table 5: All possible columns for a three-class problem. Note that the last four columns are complements of the first four and that the first column does not discriminate among any of the classes.

| Class | Code Word | | | | | | | |
|---|---|---|---|---|---|---|---|---|
| | $f_0$ | $f_1$ | $f_2$ | $f_3$ | $f_4$ | $f_5$ | $f_6$ | $f_7$ |
| $c_0$ | 0 | 0 | 0 | 0 | 1 | 1 | 1 | 1 |
| $c_1$ | 0 | 0 | 1 | 1 | 0 | 0 | 1 | 1 |
| $c_2$ | 0 | 1 | 0 | 1 | 0 | 1 | 0 | 1 |

Unless the number of classes is at least five, it is difficult to satisfy both of these properties. For example, when the number of classes is three, there are only $2^3 = 8$ possible columns (see Table 5). Of these, half are complements of the other half. So this leaves us with only four possible columns. One of these will be either all zeroes or all ones, which will make it useless for discriminating among the rows. The result is that we are left with only three possible columns, which is exactly what the one-per-class encoding provides.

In general, if there are $k$ classes, there will be at most $2^{k-1} - 1$ usable columns after removing complements and the all-zeros or all-ones column. For four classes, we get a seven-column code with minimum inter-row Hamming distance 4. For five classes, we get a 15-column code, and so on.

We have employed four methods for constructing good error-correcting output codes in this paper: (a) an exhaustive technique, (b) a method that selects columns from an exhaustive code, (c) a method based on a randomized hill-climbing algorithm, and (d) BCH codes. The choice of which method to use is based on the number of classes, $k$. Finding a single method suitable for all values of $k$ is an open research problem. We describe each of our four methods in turn.

### 2.3.1 EXHAUSTIVE CODES

When $3 \leq k \leq 7$, we construct a code of length $2^{k-1} - 1$ as follows. Row 1 is all ones. Row 2 consists of $2^{k-2}$ zeroes followed by $2^{k-2} - 1$ ones. Row 3 consists of $2^{k-3}$ zeroes, followed by $2^{k-3}$ ones, followed by $2^{k-3}$ zeroes, followed by $2^{k-3} - 1$ ones. In row $i$, there are alternating runs of $2^{k-i}$ zeroes and ones. Table 6 shows the exhaustive code for a five-class problem. This code has inter-row Hamming distance 8; no columns are identical or complementary.

### 2.3.2 COLUMN SELECTION FROM EXHAUSTIVE CODES

When $8 \leq k \leq 11$, we construct an exhaustive code and then select a good subset of its columns. We formulate this as a propositional satisfiability problem and apply the GSAT algorithm (Selman, Levesque, & Mitchell, 1992) to attempt a solution. A solution is required to include exactly $L$ columns (the desired length of the code) while ensuring that the Hamming distance between every two columns is between $d$ and $L - d$, for some chosen value of $d$. Each column is represented by a boolean variable. A pairwise mutual exclusion constraint is placed between any two columns that violate the column separation condition. To support these constraints, we extended GSAT to support mutual exclusion and "m-of-n" constraints efficiently.

Table 6: Exhaustive code for $k$=5.

| Row | Column | | | | | | | | | | | | | | |
|---|---|---|---|---|---|---|---|---|---|---|---|---|---|---|---|
| | 1 | 2 | 3 | 4 | 5 | 6 | 7 | 8 | 9 | 10 | 11 | 12 | 13 | 14 | 15 |
| 1 | 1 | 1 | 1 | 1 | 1 | 1 | 1 | 1 | 1 | 1 | 1 | 1 | 1 | 1 | 1 |
| 2 | 0 | 0 | 0 | 0 | 0 | 0 | 0 | 0 | 1 | 1 | 1 | 1 | 1 | 1 | 1 |
| 3 | 0 | 0 | 0 | 0 | 1 | 1 | 1 | 1 | 0 | 0 | 0 | 0 | 1 | 1 | 1 |
| 4 | 0 | 0 | 1 | 1 | 0 | 0 | 1 | 1 | 0 | 0 | 1 | 1 | 0 | 0 | 1 |
| 5 | 0 | 1 | 0 | 1 | 0 | 1 | 0 | 1 | 0 | 1 | 0 | 1 | 0 | 1 | 0 |

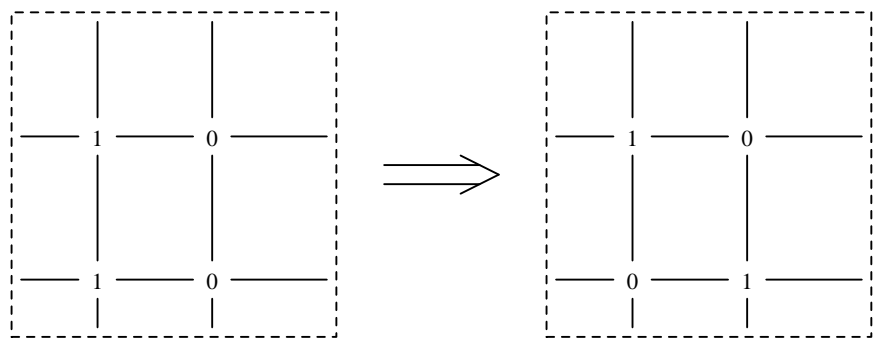

Figure 1: Hill-climbing algorithm for improving row and column separation. The two closest rows and columns are indicated by lines. Where these lines intersect, the bits in the code words are changed to improve separations as shown on the right.

### 2.3.3 Randomized Hill Climbing

For $k > 11$, we employed a random search algorithm that begins by drawing $k$ random strings of the desired length $L$. Any pair of such random strings will be separated by a Hamming distance that is binomially distributed with mean $L/2$. Hence, such randomly generated codes are generally quite good on average. To improve them, the algorithm repeatedly finds the pair of rows closest together in Hamming distance and the pair of columns that have the "most extreme" Hamming distance (i.e., either too close or too far apart). The algorithm then computes the four codeword bits where these rows and columns intersect and changes them to improve the row and column separations as shown in Figure 1. When this hill climbing procedure reaches a local maximum, the algorithm randomly chooses pairs of rows and columns and tries to improve their separations. This combined hill-climbing/random-choice procedure is able to improve the minimum Hamming distance separation quite substantially.

### 2.3.4 BCH CODES

For $k > 11$ we also applied the BCH algorithm to design codes (Bose & Ray-Chaudhuri, 1960; Hocquenghem, 1959). The BCH algorithm employs algebraic methods from Galois field theory to design nearly optimal error-correcting codes. However, there are three practical drawbacks to using this algorithm. First, published tables of the primitive polynomials required by this algorithm only produce codes up to length 64, since this is the largest word size employed in computer memories. Second, the codes do not always exhibit good column separations. Third, the number of rows in these codes is always a power of two. If the number of classes $k$ in our learning problem is not a power of two, we must shorten the code by deleting rows (and possible columns) while maintaining good row and column separations. We have experimented with various heuristic greedy algorithms for code shortening. For most of the codes used in the NETtalk, ISOLET, and Letter Recognition domains, we have used a combination of simple greedy algorithms and manual intervention to design good shortened BCH codes.

In each of the data sets that we studied, we designed a series of error-correcting codes of increasing lengths. We executed each learning algorithm for each of these codes. We stopped lengthening the codes when performance appeared to be leveling off.

## 2.4 Making Classification Decisions

Each approach to solving multiclass problems—direct multiclass, one-per-class, and error-correcting output coding—assumes a method for classifying new examples. For the C4.5 direct multiclass approach, the C4.5 system computes a class probability estimate for each new example. This estimates the probability that that example belongs to each of the $k$ classes. C4.5 then chooses the class having the highest probability as the class of the example.

For the one-per-class approach, each decision tree or neural network output unit can be viewed as computing the probability that the new example belongs to its corresponding class. The class whose decision tree or output unit gives the highest probability estimate is chosen as the predicted class. Ties are broken arbitrarily in favor of the class that comes first in the class ordering.

For the error-correcting output code approach, each decision tree or neural network output unit can be viewed as computing the probability that its corresponding bit in the codeword is one. Call these probability values $B = \langle b_1, b_2, \ldots, b_n \rangle$, where $n$ is the length of the codewords in the error-correcting code. To classify a new example, we compute the $L^1$ distance between this probability vector $B$ and each of the codewords $W_i$ ($i = 1 \ldots k$) in the error correcting code. The $L^1$ distance between $B$ and $W_i$ is defined as

$$L^1(B, W_i) = \sum_{j=0}^{L} |b_j - W_{i,j}|.$$

The class whose codeword has the smallest $L^1$ distance to $B$ is assigned as the class of the new example. Ties are broken arbitrarily in favor of the class that comes first in the class ordering.

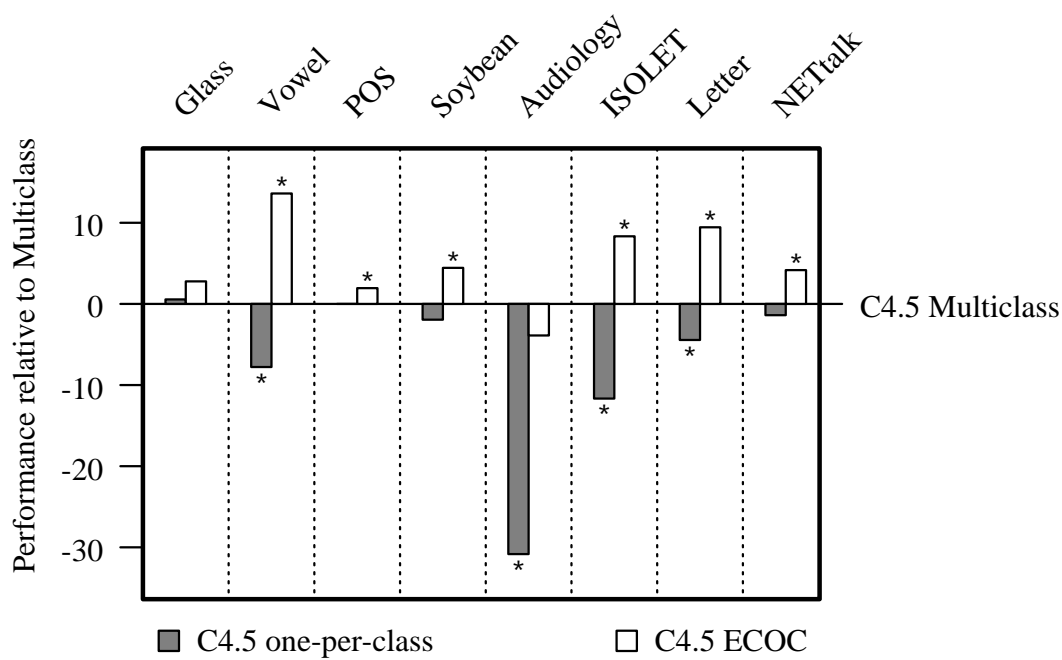

Figure 2: Performance (in percentage points) of the one-per-class and ECOC methods relative to the direct multiclass method using C4.5. Asterisk indicates difference is significant at the 0.05 level or better.

## 3. Results

We now present the results of our experiments. We begin with the results for decision trees. Then, we consider neural networks. Finally, we report the results of a series of experiments to assess the robustness of the error-correcting output code method.

### 3.1 Decision Trees

Figure 2 shows the performance of C4.5 in all eight domains. The horizontal line corresponds to the performance of the standard multiclass decision-tree algorithm. The light bar shows the performance of the one-per-class approach, and the dark bar shows the performance of the ECOC approach with the longest error-correcting code tested. Performance is displayed as the number of percentage points by which each pair of algorithms differ. An asterisk indicates that the difference is statistically significant at the $p < 0.05$ level according to the test for the difference of two proportions (using the normal approximation to the binomial distribution, see Snedecor & Cochran, 1989, p. 124).

From this figure, we can see that the one-per-class method performs significantly worse than the multiclass method in four of the eight domains and that its behavior is statistically indistinguishable in the remaining four domains. Much more encouraging is the observation that the error-correcting output code approach is significantly superior to the multiclass approach in six of the eight domains and indistinguishable in the remaining two.

In the NETtalk domain, we can also consider the performance of the meaningful distributed representation developed by Sejnowski and Rosenberg. This representation gave 66.7% correct classification as compared with 68.6% for the one-per-class configuration, 70.0% for the direct-multiclass configuration, and 74.3% for the ECOC configuration. The differences in each of these figures are statistically significant at the 0.05 level or better except that the one-per-class and direct-multiclass configurations are not statistically distinguishable.

## 3.2 Backpropagation

Figure 3 shows the results for backpropagation in five of the most challenging domains. The horizontal line corresponds to the performance of the one-per-class encoding for this method. The bars show the number of percentage points by which the error-correcting output coding representation outperforms the one-per-class representation. In four of the five domains, the ECOC encoding is superior; the differences are statistically significant in the Vowel, NETtalk, and ISOLET domains.[2]

In the letter recognition domain, we encountered great difficulty in successfully training networks using the CNAPS machine, particularly for the ECOC configuration. Experiments showed that the problem arose from the fact that the CNAPS implementation of backpropagation employs a fixed learning rate. We therefore switched to the much slower `opt` program, which chooses the learning rate adaptively via conjugate-gradient line searches. This behaved better for both the one-per-class and ECOC configurations.

We also had some difficulty training ISOLET in the ECOC configuration on large networks (182 units), even with the `opt` program. Some sets of initial random weights led to local minima and poor performance on the validation set.

In the NETtalk task, we can again compare the performance of the Sejnowski-Rosenberg distributed encoding to the one-per-class and ECOC encodings. The distributed encoding yielded a performance of 71.5% correct, compared to 72.9% for the one-per-class encoding, and 74.9% for the ECOC encoding. The difference between the distributed encoding and the one-per-class encoding is not statistically significant. From these results and the previous results for C4.5, we can conclude that the distributed encoding has no advantages over the one-per-class and ECOC encoding in this domain.

## 3.3 Robustness

These results show that the ECOC approach performs as well as, and often better than, the alternative approaches. However, there are several important questions that must be answered before we can recommend the ECOC approach without reservation:

- Do the results hold for small samples? We have found that decision trees learned using error-correcting codes are much larger than those learned using the one-per-class or multiclass approaches. This suggests that with small sample sizes, the ECOC method may not perform as well, since complex trees usually require more data to be learned reliably. On the other hand, the experiments described above covered a wide range of training set sizes, which suggests that the results may not depend on having a large training set.

---

2. The difference for ISOLET is only detectable using a test for paired differences of proportions. See Snedecor & Cochran (1989, p. 122.).

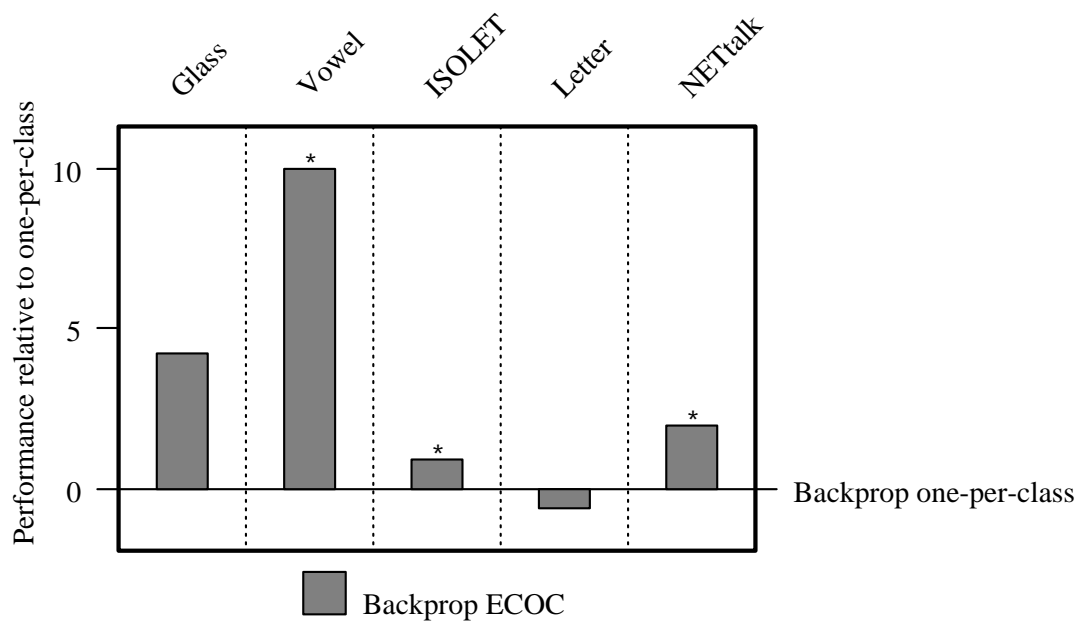

Figure 3: Performance of the ECOC method relative to the one-per-class using backprop-agation. Asterisk indicates difference is significant at the 0.05 level or better.

- Do the results depend on the particular assignment of codewords to classes? The codewords were assigned to the classes arbitrarily in the experiments reported above, which suggests that the particular assignment may not be important. However, some assignments might still be much better than others.

- Do the results depend on whether pruning techniques are applied to the decision-tree algorithms? Pruning methods have been shown to improve the performance of multiclass C4.5 in many domains.

- Can the ECOC approach provide class probability estimates? Both C4.5 and back-propagation can be configured to provide estimates of the probability that a test example belongs to each of the $k$ possible classes. Can the ECOC approach do this as well?

### 3.3.1 SMALL SAMPLE PERFORMANCE

As we have noted, we became concerned about the small sample performance of the ECOC method when we noticed that the ECOC method always requires much larger decision trees than the OPC method. Table 7 compares the sizes of the decision trees learned by C4.5 under the multiclass, one-per-class, and ECOC configurations for the letter recognition task and the NETtalk task. For the OPC and ECOC configurations, the tables show the average number of leaves in the trees learned for each bit position of the output representation. For letter recognition, the trees learned for a 207-bit ECOC are more than six times larger than those learned for the one-per-class representation. For the phoneme classification part of NETtalk, the ECOC trees are 14 times larger than the OPC trees. Another way to compare the sizes of the trees is to consider the total number of leaves in the trees. The tables clearly show that the multiclass approach requires much less memory (many fewer total leaves) than either the OPC or the ECOC approaches.

Table 7: Size of decision trees learned by C4.5 for the letter recognition task and the NETtalk task.

| Letter Recognition | Leaves per bit | Total leaves |
|---|---|---|
| Multiclass | | 2353 |
| One-per-class | 242 | 6292 |
| 207-bit ECOC | 1606 | 332383 |

| NETtalk | Leaves per bit | | Total leaves | |
|---|---|---|---|---|
| | phoneme | stress | phoneme | stress |
| Multiclass | | | 1425 | 1567 |
| One-per-Class | 61 | 600 | 3320 | 3602 |
| 159-bit ECOC | 901 | 911 | 114469 | 29140 |

With backpropagation, it is more difficult to determine the amount of "network resources" that are consumed in training the network. One approach is to compare the number of hidden units that give the best generalization performance. In the ISOLET task, for example, the one-per-class encoding attains peak validation set performance with a 78-hidden-unit network, whereas the 30-bit error-correcting encoding attained peak validation set performance with a 156-hidden-unit network. In the letter recognition task, peak performance for the one-per-class encoding was obtained with a network of 120-hidden units compared to 200 hidden units for a 62-bit error-correcting output code.

From the decision tree and neural network sizes, we can see that, in general, the error-correcting output representation requires more complex hypotheses than the one-per-class representation. From learning theory and statistics, we known that complex hypotheses typically require more training data than simple ones. On this basis, one might expect that the performance of the ECOC method would be very poor with small training sets. To test this prediction, we measured performance as a function of training set size in two of the larger domains: NETtalk and letter recognition.

Figure 4 presents learning curves for C4.5 on the NETtalk and letter recognition tasks, which show accuracy for a series of progressively larger training sets. From the figure it is clear that the 61-bit error-correcting code consistently outperforms the other two configurations by a nearly constant margin. Figure 5 shows corresponding results for backpropagation on the NETtalk and letter recognition tasks. On the NETtalk task, the results are the same: sample size has no apparent influence on the benefits of error-correcting output coding. However, for the letter-recognition task, there appears to be an interaction.

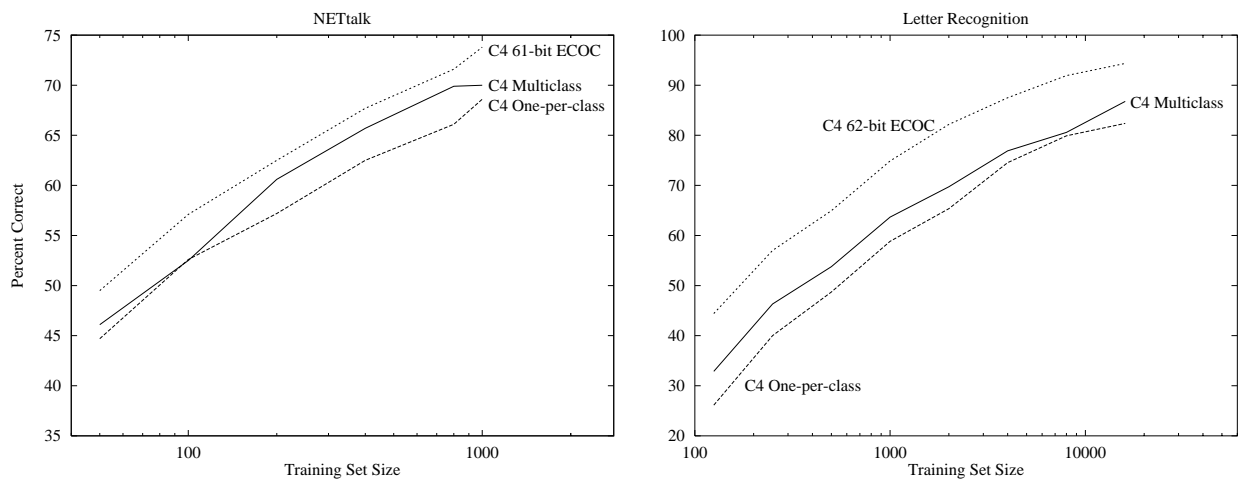

Figure 4: Accuracy of C4.5 in the multiclass, one-per-class, and error-correcting output coding configurations for increasing training set sizes in the NETtalk and letter recognition tasks. Note that the horizontal axis is plotted on a logarithmic scale.

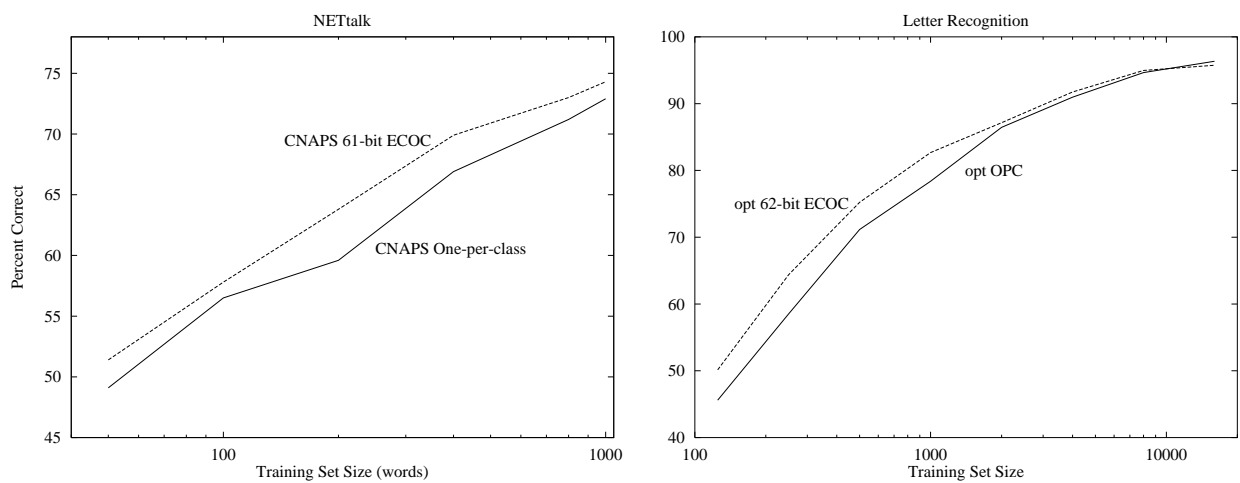

Figure 5: Accuracy of backpropagation in the one-per-class and error-correcting output coding configurations for increasing training set sizes on the NETtalk and letter recognition tasks.

Error-correcting output coding works best for small training sets, where there is a statistically significant benefit. With the largest training set—16,000 examples—the one-per-class method very slightly outperforms the ECOC method.

From these experiments, we conclude that error-correcting output coding works very well with small samples, despite the increased size of the decision trees and the increased complexity of training neural networks. Indeed, with backpropagation on the letter recognition task, error-correcting output coding worked better for small samples than it did for large ones. This effect suggests that ECOC works by reducing the variance of the learning algorithm. For small samples, the variance is higher, so ECOC can provide more benefit.

Table 8: Five random assignments of codewords to classes for the NETtalk task. Each column shows the percentage of letters correctly classified by C4.5 decision trees.

| | | 61-Bit Error-Correcting Code Replications | | | | |
|---|---|---|---|---|---|---|
| Multiclass | One-per-class | a | b | c | d | e |
| 70.0 | 68.6 | 73.8 | 73.6 | 73.5 | 73.8 | 73.3 |

### 3.3.2 ASSIGNMENT OF CODEWORDS TO CLASSES

In all of the results reported thus far, the codewords in the error-correcting code have been arbitrarily assigned to the classes of the learning task. We conducted a series of experiments in the NETtalk domain with C4.5 to determine whether randomly reassigning the codewords to the classes had any effect on the success of ECOC. Table 8 shows the results of five random assignments of codewords to classes. There is no statistically significant variation in the performance of the different random assignments. This is consistent with similar experiments reported in Bakiri (1991).

### 3.3.3 EFFECT OF TREE PRUNING

Pruning of decision trees is an important technique for preventing overfitting. However, the merit of pruning varies from one domain to another. Figure 6 shows the change in performance due to pruning in each of the eight domains and for each of the three configurations studied in this paper: multiclass, one-per-class, and error-correcting output coding.

From the figure, we see that in most cases pruning makes no statistically significant difference in performance (aside from the POS task, where it decreases the performance of all three configurations). Aside from POS, only one of the statistically significant changes involves the ECOC configuration, while two affect the one-per-class configuration, and one affects the multiclass configuration. These data suggest that pruning only occasionally has a major effect on *any* of these configurations. There is no evidence to suggest that pruning affects one configuration more than another.

### 3.3.4 CLASS PROBABILITY ESTIMATES

In many applications, it is important to have a classifier that cannot only classify new cases well but also estimate the probability that a new case belongs to each of the $k$ classes. For example, in medical diagnosis, a simple classifier might classify a patient as "healthy" because, given the input features, that is the most likely class. However, if there is a non-zero probability that the patient has a life-threatening disease, the right choice for the physician may still be to prescribe a therapy for that disease.

A more mundane example involves automated reading of handwritten postal codes on envelopes. If the classifier is very confident of its classification (i.e., because the estimated probabilities are very strong), then it can proceed to route the envelope. However, if it is uncertain, then the envelope should be "rejected", and sent to a human being who can attempt to read the postal code and process the envelope (Wilkinson, Geist, Janet, et al., 1992).

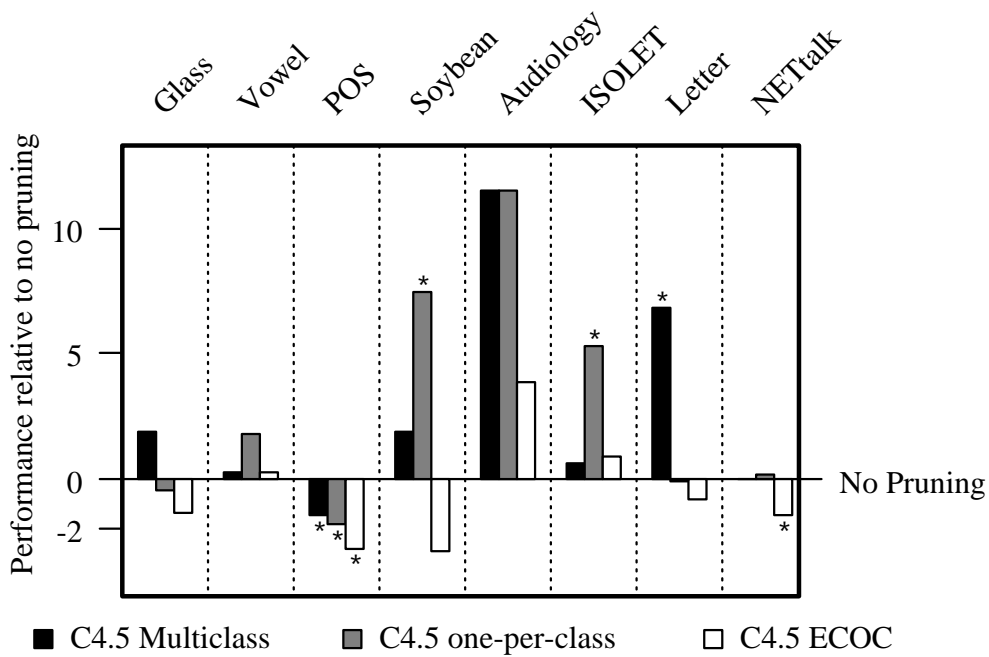

Figure 6: Change in percentage points of the performance of C4.5 with and without pruning in three configurations. Horizontal line indicates performance with no pruning. Asterisk indicates that the difference is significant at the 0.05 level or better.

One way to assess the quality of the class probability estimates of a classifier is to compute a "rejection curve". When the learning algorithm classifies a new case, we require it to also output a "confidence" level. Then we plot a curve showing the percentage of correctly classified test cases whose confidence level exceeds a given value. A rejection curve that increases smoothly demonstrates that the confidence level produced by the algorithm can be transformed into an accurate probability measure.

For one-per-class neural networks, many researchers have found that the difference in activity between the class with the highest activity and the class with the second-highest activity is a good measure of confidence (e.g., LeCun et al., 1989). If this difference is large, then the chosen class is clearly much better than the others. If the difference is small, then the chosen class is nearly tied with another class. This same measure can be applied to the class probability estimates produced by C4.5.

An analogous measure of confidence for error-correcting output codes can be computed from the $L^1$ distance between the vector $B$ of output probabilities for each bit and the codewords of each of the classes. Specifically, we employ the difference between the $L^1$ distance to the second-nearest codeword and the $L^1$ distance to the nearest codeword as our confidence measure. If this difference is large, an algorithm can be quite confident of its classification decision. If the difference is small, the algorithm is not confident.

Figure 7 compares the rejection curves for various configurations of C4.5 and backprop-agation on the NETtalk task. These curves are constructed by first running all of the test examples through the learned decision trees and computing the predicted class of each example and the confidence value for that prediction. To generate each point along the curve, a value is chosen for a parameter $\theta$, which defines the minimum required confidence. The classified test examples are then processed to determine the percentage of test examples whose confidence level is less than $\theta$ (these are "rejected") and the percentage of the remaining examples that are correctly classified. The value of $\theta$ is progressively incremented (starting at 0) until all test examples are rejected.

The lower left portion of the curve shows the performance of the algorithm when $\theta$ is small, so only the least confident cases are rejected. The upper right portion of the curve shows the performance when $\theta$ is large, so only the most confident cases are classified. Good class probability estimates produce a curve that rises smoothly and monotonically. A flat or decreasing region in a rejection curve reveals cases where the confidence estimate of the learning algorithm is unrelated or inversely related to the actual performance of the algorithm.

The rejection curves often terminate prior to rejecting 100% of the examples. This occurs when the final increment in $\theta$ causes *all* examples to be rejected. This gives some idea of the number of examples for which the algorithm was highly confident of its classifications. If the curve terminates early, this shows that there were very few examples that the algorithm could confidently classify.

In Figure 7, we see that—with the exception of the Multiclass configuration—the rejection curves for all of the various configurations of C4.5 increase fairly smoothly, so all of them are producing acceptable confidence estimates. The two error-correcting configurations have smooth curves that remain above all of the other configurations. This shows that the performance advantage of error-correcting output coding is maintained at all confidence levels—ECOC improves classification decisions on all examples, not just the borderline ones.

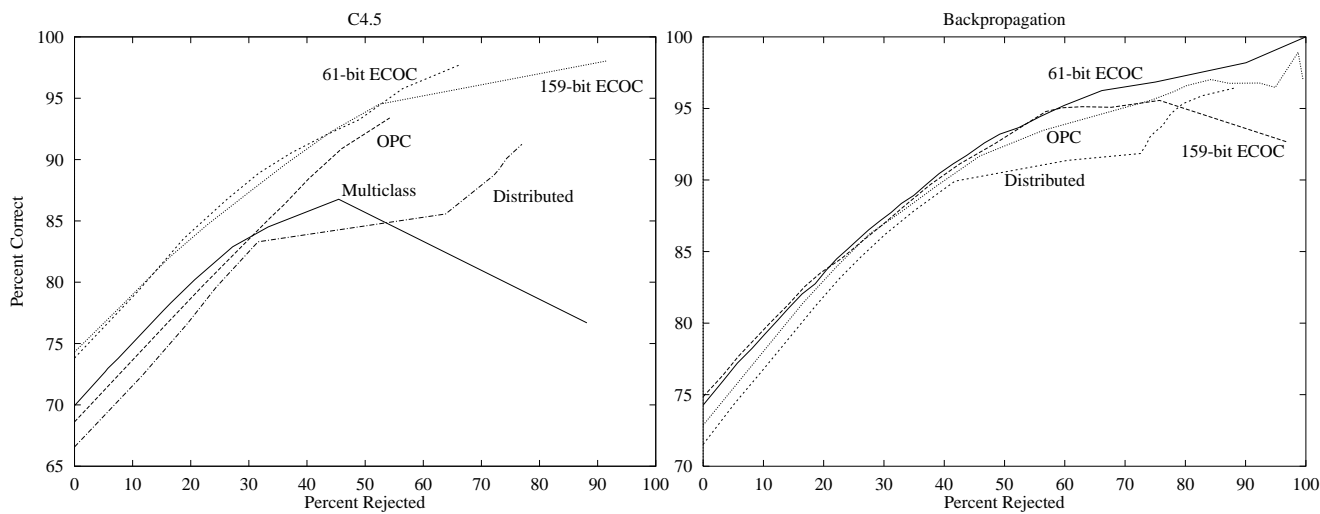

Figure 7: Rejection curves for various configurations of C4.5 and backpropagation on the NETtalk task. The "Distributed" curve plots the behavior of the Sejnowski-Rosenberg distributed representation.

Similar behavior is seen in the rejection curves for backpropagation. Again all configurations of backpropagation give fairly smooth rejection curves. However, note that the 159-bit code actually decreases at high rejection rates. By contrast, the 61-bit code gives a monotonic curve that eventually reaches 100%. We have seen this behavior in several of the cases we have studied: extremely long error-correcting codes are usually the best method at low rejection rates, but at high rejection rates, codes of "intermediate" length (typically 60-80 bits) behave better. We have no explanation for this behavior.

Figure 8 compares the rejection curves for various configurations of C4.5 and backpropagation on the ISOLET task. Here we see that the ECOC approach is markedly superior to either the one-per-class or multiclass approaches. This figure illustrates another phenomenon we have frequently observed: the curve for multiclass C4.5 becomes quite flat and terminates very early, and the one-per-class curve eventually surpasses it. This suggests that there may be opportunities to improve the class probability estimates produced by C4.5 on multiclass trees. (Note that we employed "softened thresholds" in these experiments.) In the backpropagation rejection curves, the ECOC approach consistently outperforms the one-per-class approach until both are very close to 100% correct. Note that both configurations of backpropagation can confidently classify more than 50% of the test examples with 100% accuracy.

From these graphs, it is clear that the error-correcting approach (with codes of intermediate length) can provide confidence estimates that are at least as good as those provided by the standard approaches to multiclass problems.

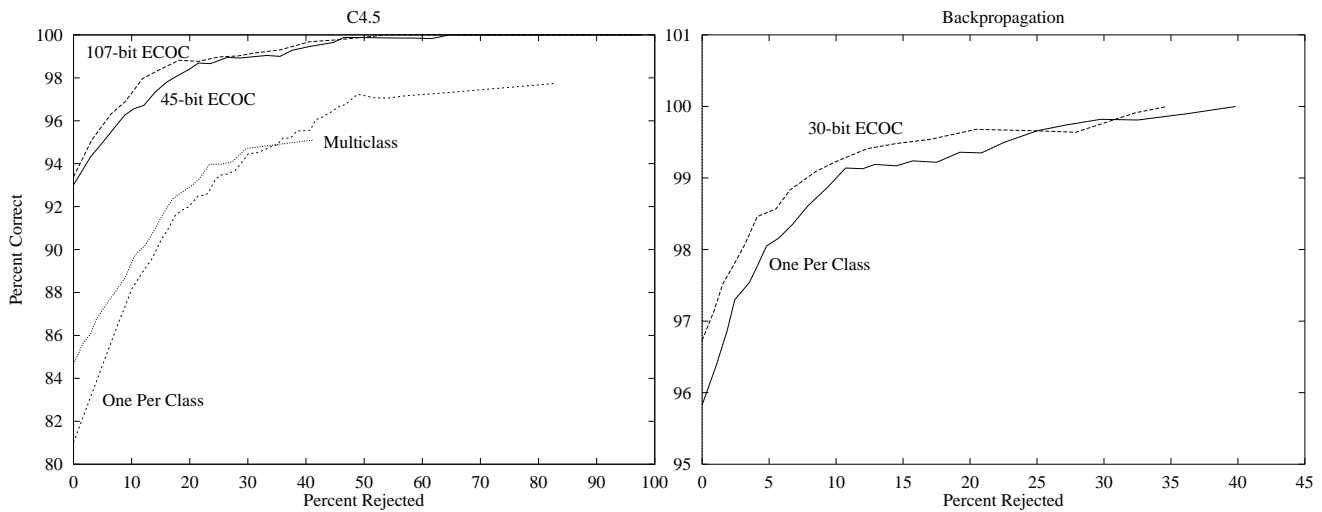

Figure 8: Rejection curves for various configurations of C4.5 and backpropagation on the ISOLET task.

## 4. Conclusions

In this paper, we experimentally compared four approaches to multiclass learning problems: multiclass decision trees, the one-per-class (OPC) approach, the meaningful distributed output approach, and the error-correcting output coding (ECOC) approach. The results clearly show that the ECOC approach is superior to the other three approaches. The improvements provided by the ECOC approach can be quite substantial: improvements on the order of ten percentage points were observed in several domains. Statistically significant improvements were observed in six of eight domains with decision trees and three of five domains with backpropagation.

The improvements were also robust:

- ECOC improves both decision trees and neural networks;

- ECOC provides improvements even with very small sample sizes; and

- The improvements do not depend on the particular assignment of codewords to classes.

The error-correcting approach can also provide estimates of the confidence of classification decisions that are at least as accurate as those provided by existing methods.

There are some additional costs to employing error-correcting output codes. Decision trees learned using ECOC are generally much larger and more complex than trees constructed using the one-per-class or multiclass approaches. Neural networks learned using ECOC often require more hidden units and longer and more careful training to obtain the improved performance (see Section 3.2). These factors may argue against using error-correcting output coding in some domains. For example, in domains where it is important for humans to understand and interpret the induced decision trees, ECOC methods are not appropriate, because they produce such complex trees. In domains where training must be rapid and completely autonomous, ECOC methods with backpropagation cannot be recommended, because of the potential for encountering difficulties during training.

Finally, we found that error-correcting codes of intermediate length tend to give better confidence estimates than very long error-correcting codes, even though the very long codes give the best generalization performance.

There are many open problems that require further research. First and foremost, it is important to obtain a deeper understanding of why the ECOC method works. If we assume that each of the learned hypotheses makes classification errors independently, then coding theory provides the explanation: individual errors can be corrected because the codewords are "far apart" in the output space. However, because each of the hypotheses is learned using the same algorithm on the same training data, we would expect that the errors made by individual hypotheses would be highly correlated, and such errors cannot be corrected by an error-correcting code. So the key open problem is to understand why the classification errors at different bit positions are fairly independent. How does the error-correcting output code result in this independence?

A closely related open problem concerns the relationship between the ECOC approach and various "ensemble", "committee", and "boosting" methods (Perrone & Cooper, 1993; Schapire, 1990; Freund, 1992). These methods construct multiple hypotheses which then "vote" to determine the classification of an example. An error-correcting code can also be viewed as a very compact form of voting in which a certain number of incorrect votes can be corrected. An interesting difference between standard ensemble methods and the ECOC approach is that in the ensemble methods, each hypothesis is attempting to predict the same function, whereas in the ECOC approach, each hypothesis predicts a different function. This may reduce the correlations between the hypotheses and make them more effective "voters." Much more work is needed to explore this relationship.

Another open question concerns the relationship between the ECOC approach and the flexible discriminant analysis technique of Hastie, Tibshirani, and Buja (In Press). Their method first employs the one-per-class approach (e.g., with neural networks) and then applies a kind of discriminant analysis to the outputs. This discriminant analysis maps the outputs into a $k-1$ dimensional space such that each class has a defined "center point". New cases are classified by mapping them into this space and then finding the nearest "center point" and its class. These center points are similar to our codewords but in a continuous space of dimension $k-1$. It may be that the ECOC method is a kind of randomized, higher-dimensional variant of this approach.

Finally, the ECOC approach shows promise of scaling neural networks to very large classification problems (with hundreds or thousands of classes) much better than the one-per-class method. This is because a good error-correcting code can have a length $n$ that is much less than the total number of classes, whereas the one-per-class approach requires that there be one output unit for each class. Networks with thousands of output units would be expensive and difficult to train. Future studies should test the scaling ability of these different approaches to such large classification tasks.

## Acknowledgements

The authors thank the anonymous reviewers for their valuable suggestions which improved the presentation of the paper. The authors also thank Prasad Tadepalli for proof-reading the final manuscript. The authors gratefully acknowledge the support of the National Science Foundation under grants numbered IRI-8667316, CDA-9216172, and IRI-9204129. Bakiri also thanks Bahrain University for its support of his doctoral research.